\documentclass{article}
\usepackage[whole]{bxcjkjatype}
\usepackage[unicode]{hyperref}
\usepackage{graphicx}
\usepackage{indentfirst}
\begin{document}

\title{A filter based approach for inbetweening}
\author{Yuichi Yagi \\ DWANGO Co., Ltd. \\ yuichi\_yagi@dwango.co.jp }
\maketitle

\begin{abstract}
We present a filter based approach for inbetweening.
We train a convolutional neural network to generate intermediate frames.
This network aim to generate smooth animation of line drawings.
Our method can process scanned images directly.
Our method does not need to compute correspondence of lines and topological changes explicitly.
We experiment our method with real animation production data.
The results show that our method can generate intermediate frames partially.
\end{abstract}

\section*{概要}
本稿では畳み込みニューラルネットワーク(CNN)を用いたラスターベースの中割フィルターについて述べる．
この画像フィルターは複数枚の線画のアニメーションからより滑らかな動きのアニメーションを作ることを目的としている．
本手法はラスターベースの画像フィルターであるためスキャンした線画に対し直接の処理が可能である．
CNNを使うことによりフレーム間の線の対応関係や線のトポロジーの変化を陽に計算する必要も無い．
本手法を実際のテレビアニメーションの動画とセルを用いて実験した．
入力画像がフレーム間で十分に似ている場合には不完全であるが中割が出来ることが確認できた．

\section{関連研究}
中割に関する先行研究については文献\cite{Dalstein:2015:vac}が詳しいのでそちらを参照されたい．
自然画像に対してはフレーム予測\cite{mathieu2015deep}\cite{liu2017video}や
フレーム補間\cite{Mahajan:2009:MGP:1531326.1531348}\cite{liu2017video}が提案されている．
しかしこれらの手法が線画でも有効であるかは明らかでない．
シモセラらはCNNを用いたラフスケッチの自動線画化を提案した\cite{SimoSerraSIGGRAPH2016}．
本研究はこの方法を出発点とし中割が出来るように変更を加えたものである．
主な変更点は以下の三点である．
\begin{itemize}
\item 高解像度画像での計算負荷を減らす為のCNNの構成
\item CNNの訓練の効率向上の為の損失関数の重み付け
\item 結果画像の品質向上の為のデータ拡大(data augmentation)
\end{itemize}
次の章でこれらについて述べる．

\section{提案手法}

\subsection{CNNの構成}
本手法は低解像度ネットワークと高解像度ネットワークで構成される．
これらのネットワークは共に2枚の線画を入力し中間的な線画を1枚出力する．
低解像度ネットワークの入力は高解像度ネットワークの入力を$1/2$に縮小したものを使う．
低解像度ネットワークは文献\cite{SimoSerraSIGGRAPH2016}のモデルを基礎にしている．
高解像度ネットワークは高解像度画像での計算負荷を減らす為に使用する．

低解像度ネットワークの各層では以下のような計算を行う．
\[
y_{u,v,c}^{0,\ell+1} = b_{c}^{0,\ell} + \sum_{d_u=-1}^{1} \sum_{d_v=-1}^{1} \sum_{i=1}^{nc_{0,\ell}} W_{c,d_u,d_v,i}^{0,\ell} x_{s_\ell \cdot u+d_u,s_\ell \cdot v+d_v,i}^{0,\ell} ,
\]
\[
x_{u,v,c}^{0,\ell+1} = {\rm ReLU} \left( y_{u,v,c}^{0,\ell+1} \right) .
\]
ここで，$x_{u,v,c}^{0,\ell}$は低解像度ネットワークの第$\ell$層の出力である．
$b_{c}^{0,\ell}$と$W_{c,d_u,d_v,i}^{0,\ell}$は訓練が必要な係数である．
高解像度ネットワークの各層では以下のような計算を行う．
\[
y_{u,v,c}^{1,\ell+1} = b_{c}^{1,\ell} + \sum_{d_u=-1}^{1} \sum_{d_v=-1}^{1} \left[ \sum_{i=1}^{nc_{1,\ell}} W_{c,d_u,d_v,i}^{1,\ell} x_{s_\ell \cdot u+d_u,s_\ell \cdot v+d_v,i}^{1,\ell} + \sum_{i=1}^{nc_{0,\ell}} W_{c,d_u,d_v,i}^{1,0,\ell} x_{s_\ell \cdot u+d_u,s_\ell \cdot v+d_v,i}^{0,\ell} \right] ,
\]
\[
x_{u,v,c}^{1,\ell+1} = {\rm ReLU} \left( y_{u,v,c}^{1,\ell+1} \right) .
\]
ここで，$x_{u,v,c}^{1,\ell}$は高解像度ネットワークの第$\ell$層の出力である．
$b_{c}^{1,\ell}$と$W_{c,d_u,d_v,i}^{1,\ell}$と$W_{c,d_u,d_v,i}^{1,0,\ell}$は訓練が必要な係数である．
$nc_{0,\ell}$と$nc_{1,\ell}$はチャンネル数で$s_\ell$はstrideである．これらの値については表\ref{table:cnn}を参照されたい．
畳み込みには3x3を使用し活性化関数にはReLUを使用する．
\[
{\rm ReLU}(x) = \max(0,x).
\]
中間層で空間解像度を上げる場合にはbilinear補間を使用する．
本研究ではBatch Normalization\cite{ioffe2015batch}は使用しない．

\subsection{損失関数}
損失関数には文献\cite{SimoSerraSIGGRAPH2016}と同様に重み付き平均二乗誤差を使用する．
\[
L = \frac{1}{nb \cdot nu \cdot nv \cdot nc} \sum_{b=1}^{nb} \sum_{u=1}^{nu} \sum_{v=1}^{nv} \sum_{c=1}^{nc} w_{b,u,v,c} \cdot (x_{b,u,v,c} - t_{b,u,v,c})^2 .
\]
$x_{b,u,v,c}$はCNNの出力画像で$t_{b,u,v,c}$は目標画像である．
$nb$はバッチサイズ，$nu$と$nv$は画像サイズ，$nc$はチャンネル数である．
重み$w_{b,u,v,c}$には文献\cite{SimoSerraSIGGRAPH2016}とは違うものを使用する．
本研究では重みの値が目標画像の線の近傍で大きな値をとり線から離れた座標で小さな値をとるような重みを使用する．
これはCNNの訓練を線の近傍に集中させることを意図している．
スキャンした動画に対して処理を行う場合は重み$w_{b,u,v,c}^{scan}$を使用する．
\[
f_{b,u,v}^{scan} = 1 - \min(t_{b,u,v,1},t_{b,u,v,2},t_{b,u,v,3}),
\]
\[
w_{b,u,v,c}^{scan} = {\rm clamp} ( \sum_{d_u=-2}^{2} \sum_{d_v=-2}^{2} f_{b,u+d_u,v+d_v}^{scan}, w_{min}, w_{max} ).
\]
${\rm clamp}$関数は以下のように定義される．
\[
{\rm clamp}(x,x_{min},x_{max}) = \min(\max(x,x_{min}),x_{max}).
\]
セルに対して処理を行う場合は重み$w_{b,u,v,c}^{cell}$を使用する．
\[
f_{b,u,v}^{cell,0} = 100 \cdot ( 0.299 \cdot t_{b,u,v,1} + 0.587 \cdot t_{b,u,v,2}  + 0.114 \cdot t_{b,u,v,3} ),
\]
\[
f_{b,u+d_u,v+d_v}^{cell,1} = \left| f_{b,u-1,v}^{cell,0} + f_{b,u+1,v}^{cell,0} + f_{b,u,v-1}^{cell,0} + f_{b,u,v+1}^{cell,0} -4 f_{b,u,v}^{cell,0} \right|,
\]
\[
w_{b,u,v,c}^{cell} = {\rm clamp} ( \sum_{d_u=-2}^{2} \sum_{d_v=-2}^{2} f_{b,u+d_u,v+d_v}^{cell,1}, w_{min}, w_{max} ).
\]
本研究では$w_{min}=1/20,w_{max}=1$を使用する．
この損失関数を低解像度ネットワークの出力と高解像度ネットワークの出力の両方で計算し，それらを平均したものを目的関数として使用する．

\subsection{データ拡大(data augmentation)}
CNNの訓練時に訓練画像に対しランダムな平行移動や回転を行いデータ拡大をする．
ここで訓練画像を$f(i,x)$としデータ拡大を施した画像を$g(i,x)$とする．
$i=0,2$は入力画像を意味し$i=1$は目標画像を意味する．
$x$は2次元の空間座標で画素の幅は$1$とする．

平行移動によるデータ拡大は以下のように行う．
\[
g(i,x) = f(i,x+(i-1)\Delta), \qquad i = 0,1,2.
\]
ここで$\Delta=(dx,dy)$で$dx$,$dy$は共に区間$[-\delta,\delta]$の整数の一様乱数である．

回転によるデータ拡大は以下のように行う．
\[
g(i,x) = f(i,R_{(i-1)\theta}(x-c)+c), \qquad i = 0,1,2.
\]
$R_{\phi}$は回転角$\phi$の回転行列である．
$\theta$は区間$[-\theta_r,\theta_r]$での一様乱数である．
回転の中心座標$c$には一様乱数でランダムに選んだ画像内の点を使う．
補間にはbilinear補間を使用する．

上記のデータ拡大を行った後に画像のランダムなクロップを行う．

\subsection{訓練と画像生成}

本研究ではCNNの訓練はカットごとに行う．
訓練結果を別のカットには使用しない．
訓練画像を$f_i$とし$i=1,2,\cdots,n_f$はフレーム番号で$n_f$を総フレーム数とする．
訓練画像のペアは3枚の画像で構成され1枚目と3枚目が入力画像で2枚目が目標画像である．
本研究では以下の二種類の訓練画像のペアを使用する．
\[
(f_{i},f_{i+1},f_{i+2}), \qquad i=1,2,\cdots,n_f-2.
\]
\[
(f_{i},f_{i},f_{i}), \qquad i=1,2,\cdots,n_f.
\]
これらの訓練画像のペアに対し平行移動か回転のデータ拡大を行ったものを訓練に使用する．

CNNの訓練後の画像生成は以下のように行う．
\[
f_{i+1/2} = {\rm CNN}(f_{i},f_{i+1}), \qquad i=1,2,\cdots,n_f-1.
\]
\[
f_{i+1/4} = {\rm CNN}(f_{i},f_{i+1/2}), \qquad i=1,2,\cdots,n_f-1.
\]
\[
f_{i+3/4} = {\rm CNN}(f_{i+1/2},f_{i+1}), \qquad i=1,2,\cdots,n_f-1.
\]

\section{結果}
提案手法をテレビアニメーションのアイドル事変\cite{idol}の動画とセルを用いて実験した．
実装にはPythonとTensorFlow\cite{tensorflow2015-whitepaper}を使い，計算には GeForce GTX TITAN X を使用した．
CNNの係数の初期化には文献\cite{He_2015_ICCV}の方法を使用し，最適化法にはAdam\cite{kingma2014adam}を使用した．
低解像度ネットワークの係数と高解像度ネットワークの係数は同時に訓練した．
CNNの各層で畳み込みを行う前にzero paddingを行った．

\subsection{実験}
実験には表\ref{table:cut}のカットを使用した．
01\_304についてはAセルとBセルを合成したものを使用した．
05\_226と05\_227と05\_228は約16枚/秒のカットで枚数の多いカットである．
これらのカットに対し計算を行い枚数を約4倍にした．
計算結果については補足動画\footnote{ \url{https://youtu.be/_RM1zUrY1AQ} }を参照されたい．
補足動画のアニメーションのタイミングは本来のタイムシートのタイミングとは異なるので注意されたい．
訓練の条件と実行時間については表\ref{table:result}と表\ref{table:result_cell}と表\ref{table:result2}を参照されたい．
訓練後の画像生成には1分から2分の時間が掛かった．

\subsubsection{動画}
A4サイズの動画用紙を300dpiでスキャンし位置合わせをした後に$1/2$に縮小した画像を使用した．
画像サイズは$1754 \times 1240$である．
平行移動によるデータ拡大は$\delta=40$で行った．
回転によるデータ拡大は使用していない．

\subsubsection{セル}
彩色されたデータに対しても実験をした．実験に使用した画像のサイズは$1686 \times 1068$である．
平行移動によるデータ拡大は$\delta=40$で行った．
回転によるデータ拡大は使用していない．

\subsubsection{動画2}
ここでは回転によるデータ拡大を使用した．$\theta_r=20^\circ$で計算を行った．
平行移動によるデータ拡大は$\delta=40$で行った．
ここでは低解像度ネットワークの中間層のチャンネル数を2倍にした．

\section{今後の課題}
本研究ではデータ拡大を訓練データの線画の動きを考慮せず行ったため無駄が多いと思われる．
フレーム間の領域の対応付け\cite{zhu-2016-toontrack}などで動きを求めて，それを元にデータ拡大をすれば無駄が減ると思われる．
CNNの中間層のチャンネル数の調節も課題である．
チャンネル数が少ないと十分に訓練をしてもボケた画像しか得られず，多すぎると訓練が進まなくなる傾向がみられた．
また，CNNに大量のカットを学習させ未知のカットに対して処理をすることも今後の課題である．

\section{まとめ}
入力画像がフレーム間で十分に似ている場合ならば不完全ながらも提案手法で中割が出来ることが確認できた．
提案手法のアニメーション制作への応用については，
原画を直接中割することは無理だと思われるため動画工程の省力化には寄与しそうにない．
しかし，秒24枚以上のアニメーションを制作する場合やスローモーションを作る場合には役立つ可能性はある．

\section{謝辞}
データを提供して下さったスタジオヴォルン様とMAGES.様，アイドル事変製作委員会様に感謝致します．
また，素晴らしいデータを使用させて頂き有難う御座います．制作に携わった方々に感謝致します．

\nocite{*}
\bibliography{main}

\begin{table}
\begin{center}
\begin{tabular}{c|r|r|r|r}
$\ell$ & $nc_{0,\ell}$ & $nc_{1,\ell}$ & $s_{\ell}$ & size \\
\hline
 0 &   0 &   6 & 2 & 1    \\
 1 &   0 &   8 & 1 & 1/2  \\
 2 &   0 &   8 & 1 & 1/2  \\
 3 &   6 &   8 & 2 & 1/2  \\
 4 &  40 &   8 & 1 & 1/4  \\
 5 &  80 &  16 & 1 & 1/4  \\
 6 &  80 &  16 & 2 & 1/4  \\
 7 & 160 &  32 & 1 & 1/8  \\
 8 & 160 &  32 & 1 & 1/8  \\
 9 & 160 &  32 & 2 & 1/8  \\
10 & 160 &  32 & 1 & 1/16 \\
11 & 320 &  64 & 1 & 1/16 \\
12 & 640 & 128 & 1 & 1/16 \\
13 & 640 & 128 & 1 & 1/16 \\
14 & 640 & 128 & 1 & 1/16 \\
15 & 640 & 128 & 1 & 1/16 \\
16 & 320 &  64 & 1 & 1/16 \\
17 & 160 &  32 & 1 & 1/16 \\
18 & 160 &  32 & 1 & 1/8  \\
19 & 160 &  32 & 1 & 1/8  \\
20 &  80 &  16 & 1 & 1/8  \\
21 &  80 &  16 & 1 & 1/4  \\
22 &  80 &  16 & 1 & 1/4  \\
23 &  40 &   8 & 1 & 1/4  \\
24 &  40 &   8 & 1 & 1/2  \\
25 &  40 &   8 & 1 & 1/2  \\
26 &   3 &  24 & 1 & 1/2  \\
27 &   0 &  24 & 1 & 1    \\
28 &   0 &  24 & 1 & 1    \\
29 &   0 &   3 &   & 1    \\
\hline
\end{tabular}
\end{center}
\caption{CNNの構成}
\label{table:cnn}
\end{table}

\begin{table}
\begin{center}
\begin{tabular}{c|c|c|c}
カット & 動画 & セル & $n_f$ \\
\hline
01\_304 & \includegraphics[width=50mm]{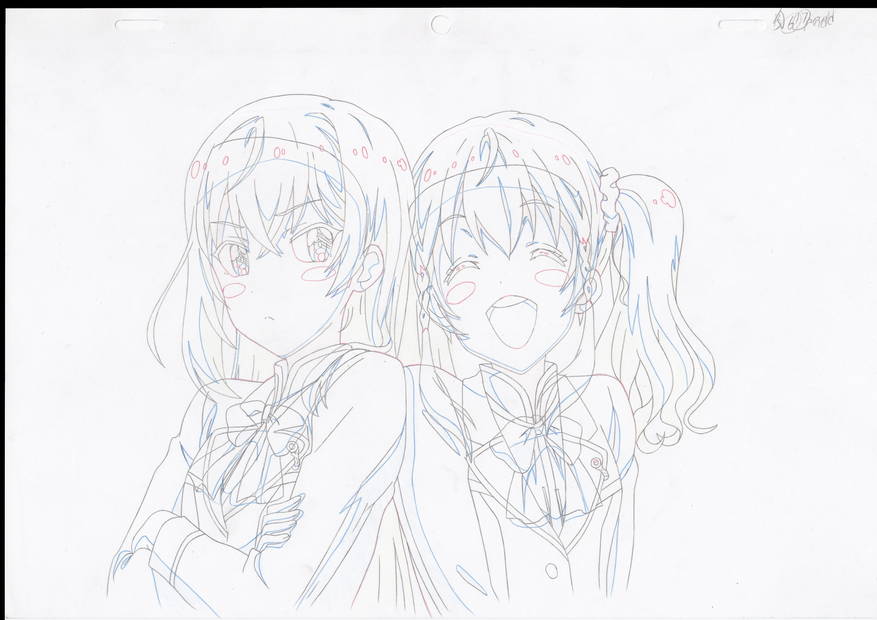} & \includegraphics[width=50mm]{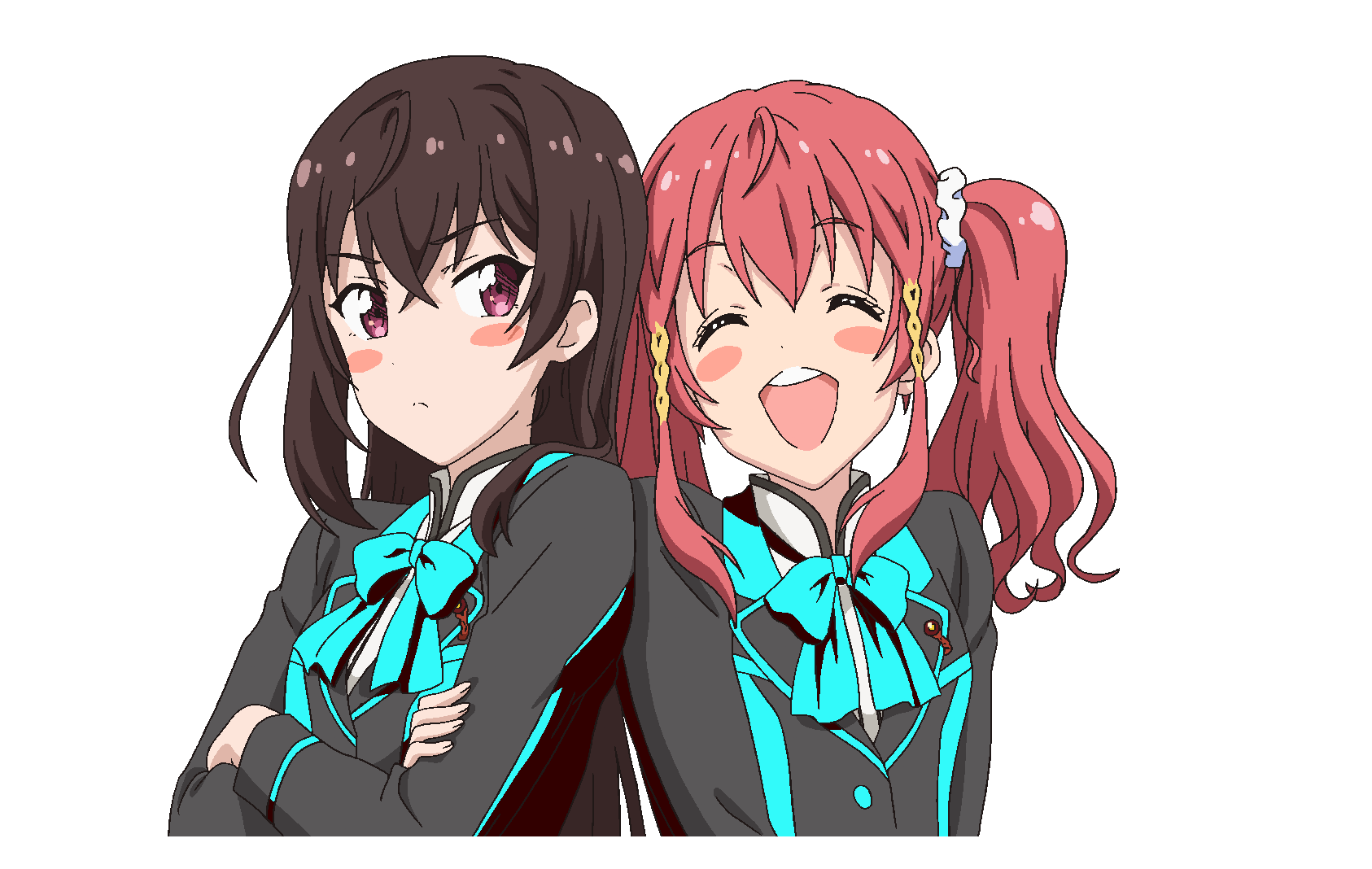} &   21   \\
\hline
05\_226 & \includegraphics[width=50mm]{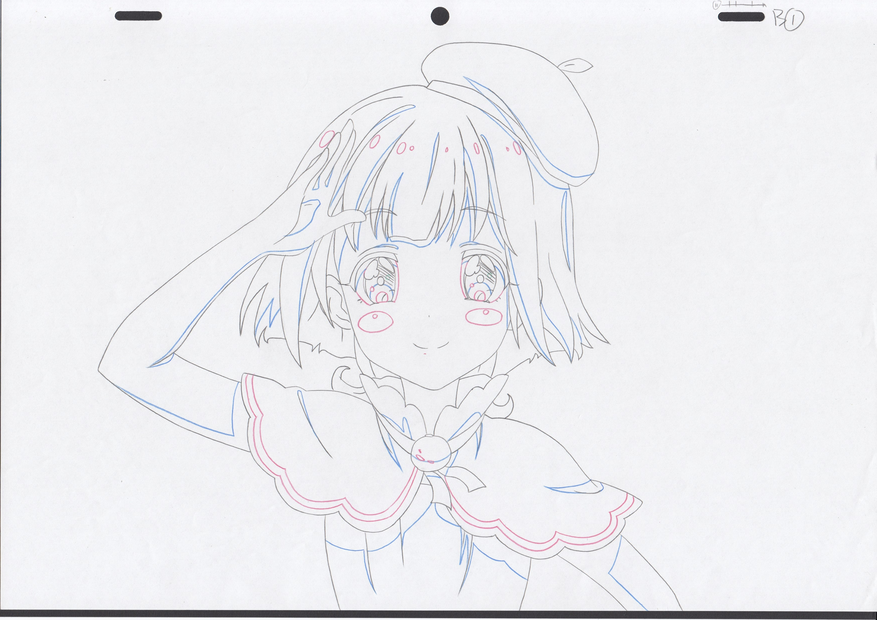} & \includegraphics[width=50mm]{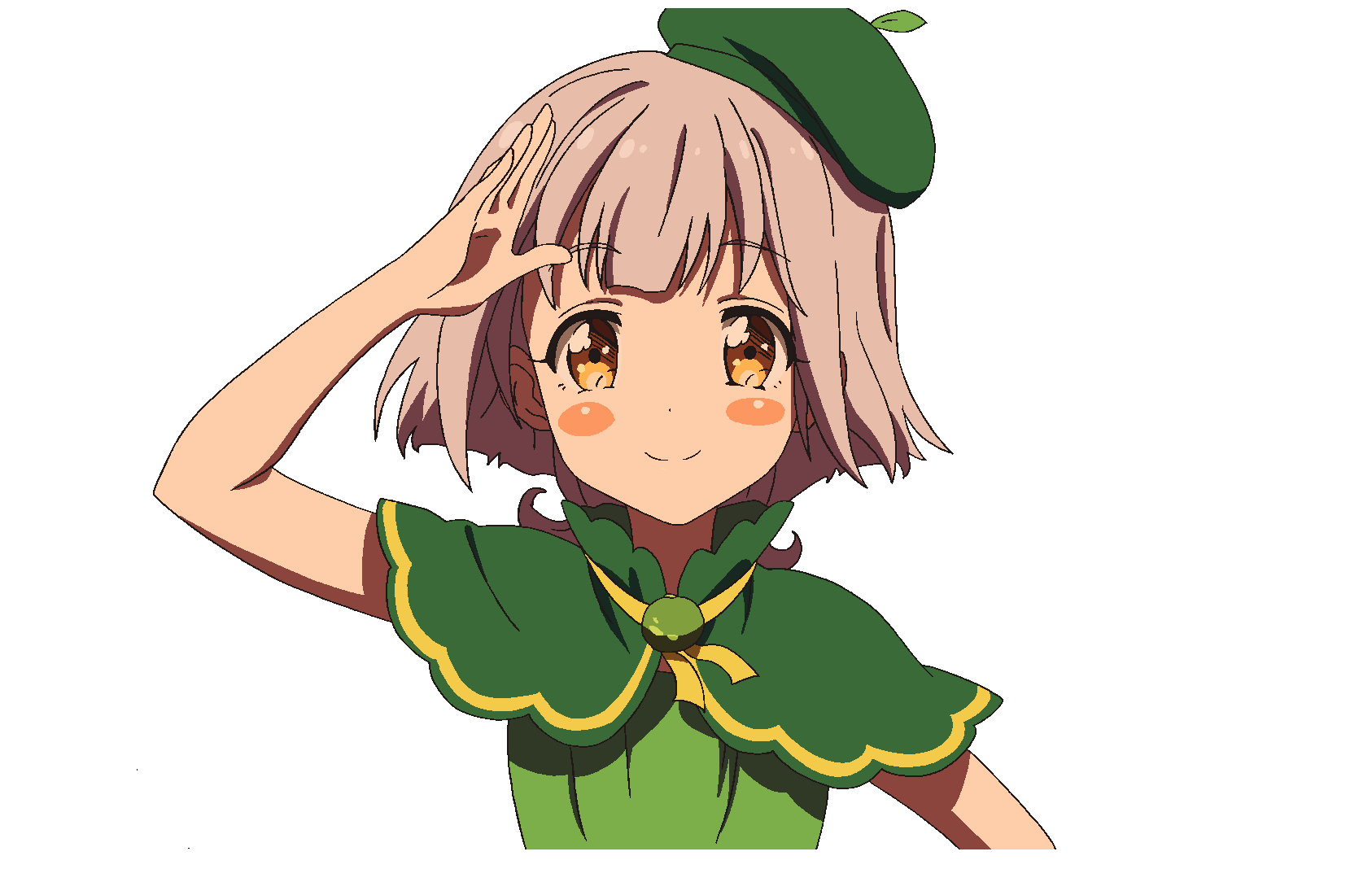} &   23   \\
\hline
05\_227 & \includegraphics[width=50mm]{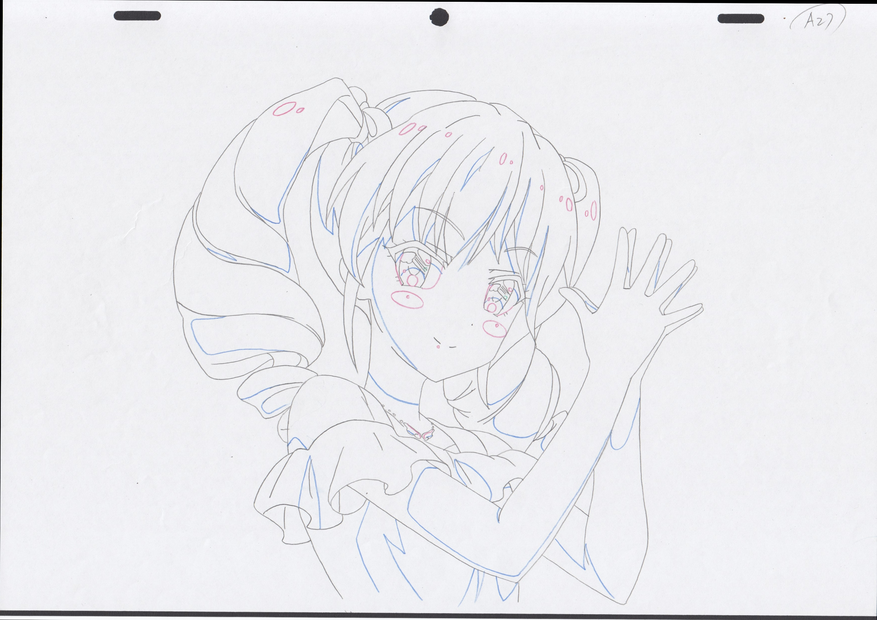} & \includegraphics[width=50mm]{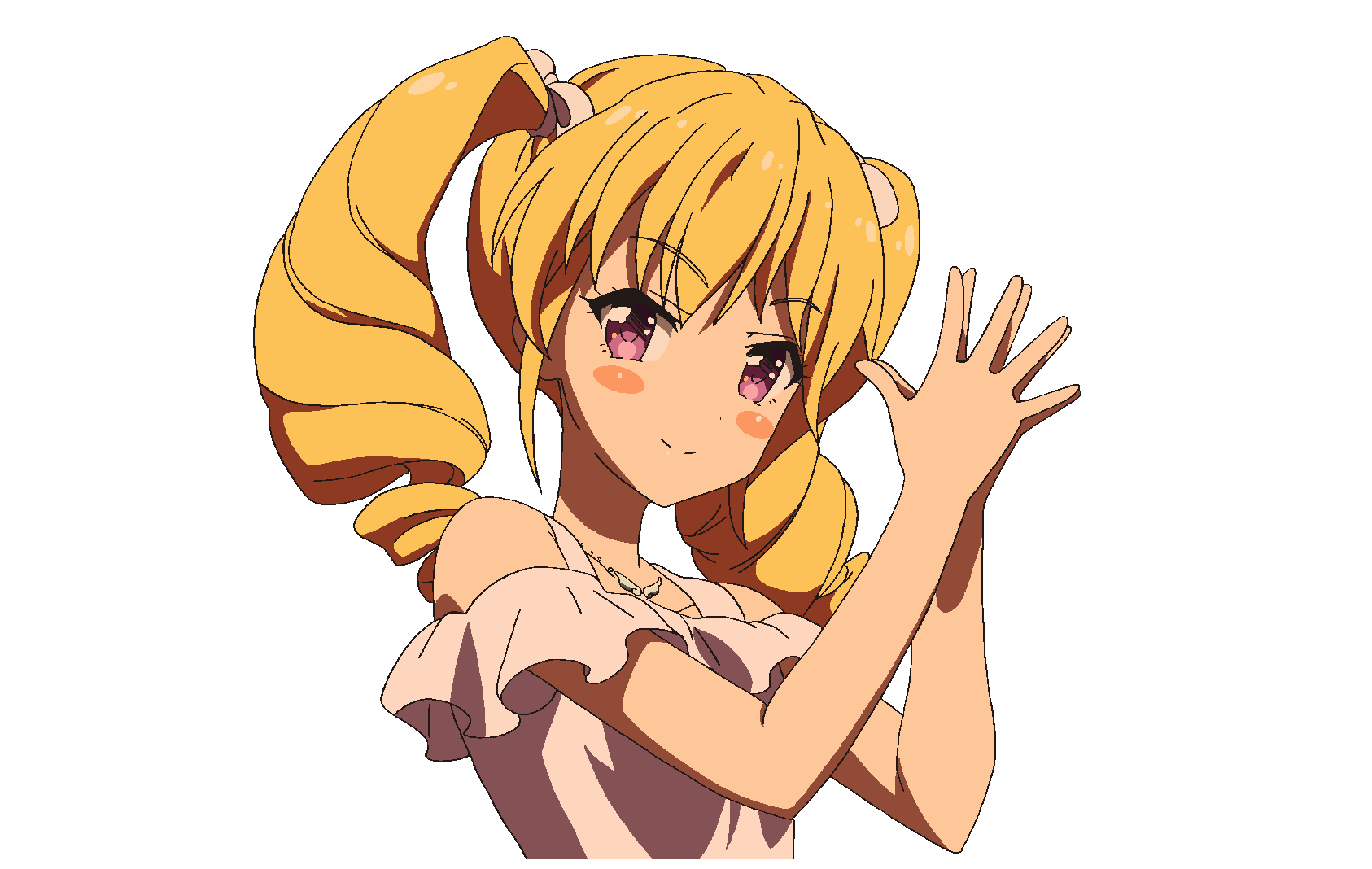} &   29   \\
\hline
05\_228 & \includegraphics[width=50mm]{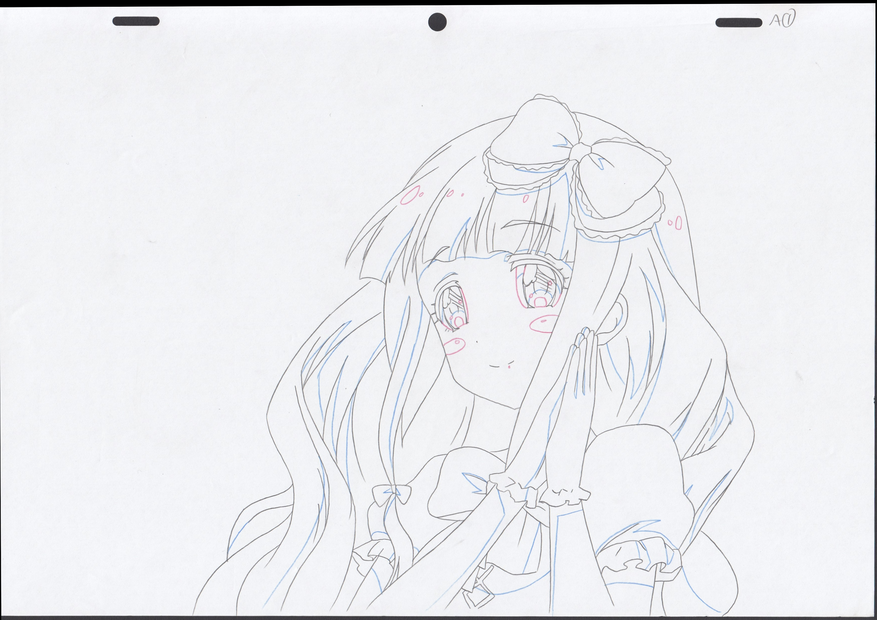} & \includegraphics[width=50mm]{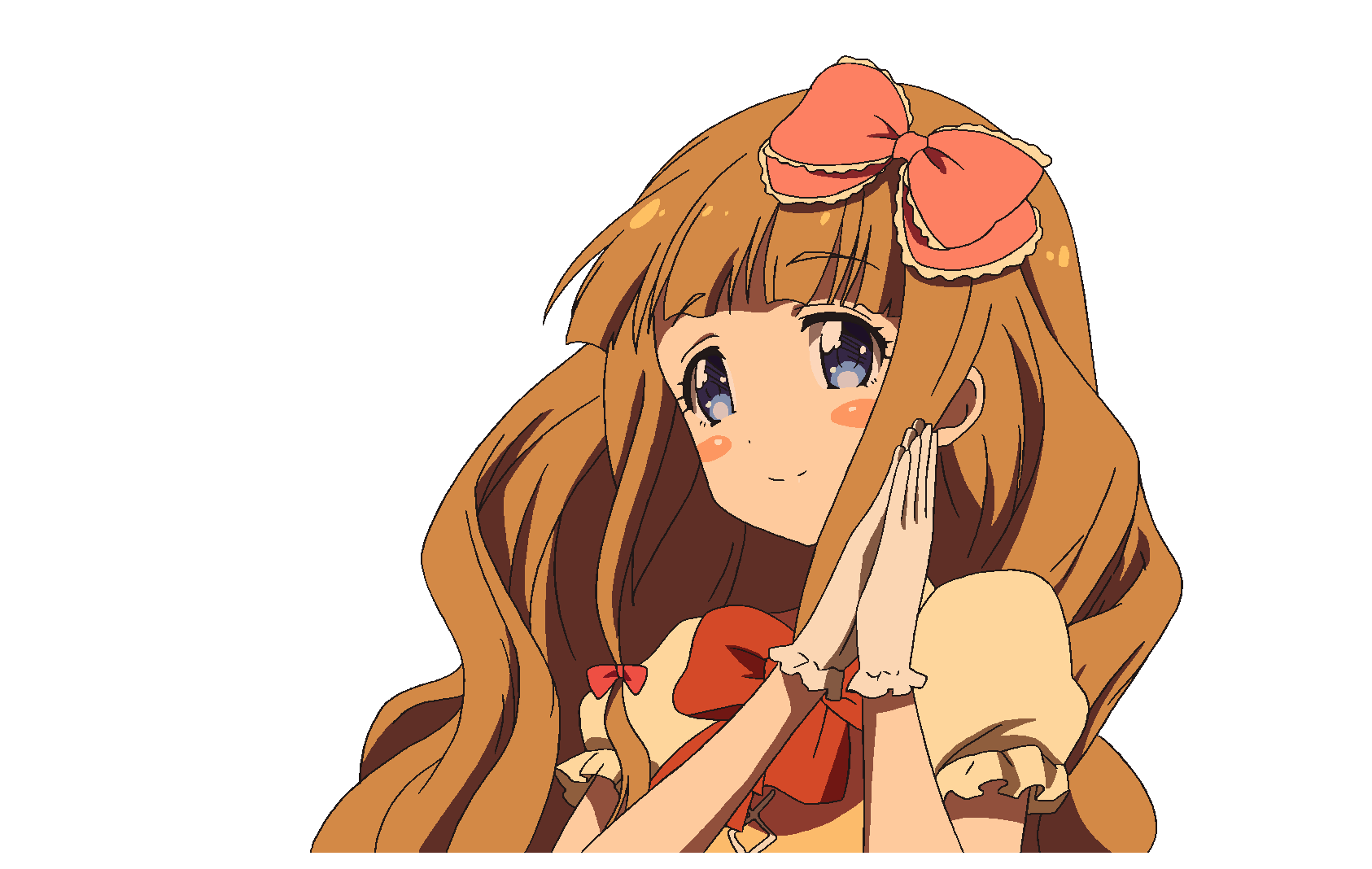} &   17   \\
\hline
\end{tabular}
\end{center}
\caption{実験に使用したカット． ©MAGES. ©アイドル事変製作委員会 }
\label{table:cut}
\end{table}
\begin{table}
\begin{center}
\begin{tabular}{c|c|r|l|r|r}
カット & 画像サイズ & バッチサイズ & 学習率 & 反復回数 & 実行時間 \\
\hline
01\_304 & $1024 \times 512$ & 11 & 0.0002 & 123000 & 71h \\
05\_226 & $1024 \times 512$ & 13 & 0.00015 & 103000 & 70h \\
05\_227 & $1024 \times 512$ & 13 & 0.0002 & 114000 & 77h \\
05\_228 & $1024 \times 512$ & 13 & 0.0002 & 135000 & 92h \\
\hline
\end{tabular}
\end{center}
\caption{訓練結果（動画）}
\label{table:result}
\end{table}

\begin{table}
\begin{center}
\begin{tabular}{c|c|r|l|r|r}
カット & 画像サイズ & バッチサイズ & 学習率 & 反復回数 & 実行時間 \\
\hline
01\_304 & $1024 \times 1024$ & 6 & 0.0002 & 155500 & 97h \\
05\_226 & $1024 \times 512$ & 13 & 0.0002 & 109000 & 73h \\
05\_227 & $1024 \times 512$ & 13 & 0.0002 & 110000 & 74h \\
05\_228 & $1024 \times 512$ & 13 & 0.0002 & 102000 & 70h \\
\hline
\end{tabular}
\end{center}
\caption{訓練結果（セル）}
\label{table:result_cell}
\end{table}

\begin{table}
\begin{center}
\begin{tabular}{c|c|r|l|r|r}
カット & 画像サイズ & バッチサイズ & 学習率 & 反復回数 & 実行時間 \\
\hline
01\_304 & $1024 \times 512$ & 8 & 0.00015 & 290000 & 313h \\
05\_226 & $1024 \times 512$ & 8 & 0.000175 & 300000 & 288h \\
05\_227 & $1024 \times 512$ & 8 & 0.00015 & 300000 & 305h \\
05\_228 & $1024 \times 512$ & 8 & 0.0002 & 300000 & 304h \\
\hline
\end{tabular}
\end{center}
\caption{訓練結果（動画2）}
\label{table:result2}
\end{table}

\end{document}